%% file: main.tex
\documentclass[letterpaper]{article} 
\usepackage[preprint]{aaai2027}  
\usepackage[hyphens]{url}  
\usepackage{graphicx} 
\usepackage{natbib}  
\usepackage{caption} 
\usepackage{algorithm}
\usepackage{algorithmic}

\usepackage{newfloat}
\usepackage{listings}
\DeclareCaptionStyle{ruled}{labelfont=normalfont,labelsep=colon,strut=off} 
\floatstyle{ruled}
\newfloat{listing}{tb}{lst}{}
\floatname{listing}{Listing}

\usepackage{booktabs}
\usepackage{amsmath}
\usepackage[capitalize]{cleveref}
    \crefname{section}{Sec.}{Secs.}
    \Crefname{section}{Sec.}{Secs.}
    \crefname{table}{Tab.}{Tabs.}
    \Crefname{table}{Tab.}{Tabs.}
    \crefname{equation}{Eq.}{Eqs.}
    \Crefname{equation}{Eq.}{Eqs.}
    \crefname{figure}{Fig.}{Figs.}
    \Crefname{figure}{Fig.}{Figs.}
\title{AdaThinkV: Adaptive Thinking for Token-Efficient Video Reasoning}
\author{
    Jingqi Tian\textsuperscript{\rm 1}\equalcontrib,
    Haoji Zhang\textsuperscript{\rm 1}\equalcontrib,
    Lin Chen\textsuperscript{\rm 2}\equalcontrib,
    Hongbo Jin\textsuperscript{\rm 3},
    Haonan Xu\textsuperscript{\rm 2},\\
    Tianrui Zhu\textsuperscript{\rm 1},
    Xingming Shui\textsuperscript{\rm 1},
    Shilin Ma\textsuperscript{\rm 1},
    Wenjing Yang\textsuperscript{\rm 2},
    Yansong Tang\textsuperscript{\rm 1}\corresponding
}
\affiliations{
    \textsuperscript{\rm 1}Tsinghua Shenzhen International Graduate School, Tsinghua University
    \textsuperscript{\rm 2}Alibaba Group
    \textsuperscript{\rm 3}Peking University\\
    \{tjq25@mails, tang.yansong@sz\}.tsinghua.edu.cn
}

\begin{document}

\maketitle

\input{sec/0_abstract}


\input{sec/1_introduction}
\input{sec/2_related_work}
\input{sec/3_method}
\input{sec/4_experiments}
\input{sec/6_conclusion}




\bibliography{main}


\end{document}

%% file: sec/0_abstract.tex

\begin{abstract}
Chain-of-thought (CoT) reasoning can improve performance on difficult video questions but often wastes decoding tokens on simple ones. We study whether a video multimodal large language model can adapt its reasoning effort to each question.
We propose AdaThinkV, an adaptive framework for video reasoning that learns whether to reason explicitly without offline difficulty labels, manually tuned confidence thresholds, or an external router.
During reinforcement learning, AdaThinkV samples matched rollouts in explicit reasoning and direct answering modes for each prompt.
ThinkGain estimates the prompt-level utility of explicit reasoning by balancing its accuracy gain against additional response length, providing supervision for both conditional response generation and autonomous mode selection.
For difficult prompts, limited rollout exploration can yield groups in which every response is unsuccessful and accuracy rewards show little variation, providing insufficient signal for learning. We therefore introduce Variance Recovery Policy Optimization (VRPO), which retains and progressively expands these groups to recover informative signals from prompts that are difficult yet solvable. At inference, AdaThinkV selects a response mode and generates the response in a single autoregressive sequence. Across a unified suite of video reasoning evaluations, AdaThinkV achieves a mean accuracy of 40.79 with an average of 257.20 output tokens, outperforming the strongest evaluated adaptive baseline by 2.98 points while using 22.7\% fewer tokens.
\end{abstract}

%% file: sec/1_introduction.tex
\begin{figure*}[!h]
\centering
\includegraphics[width=0.98\textwidth]{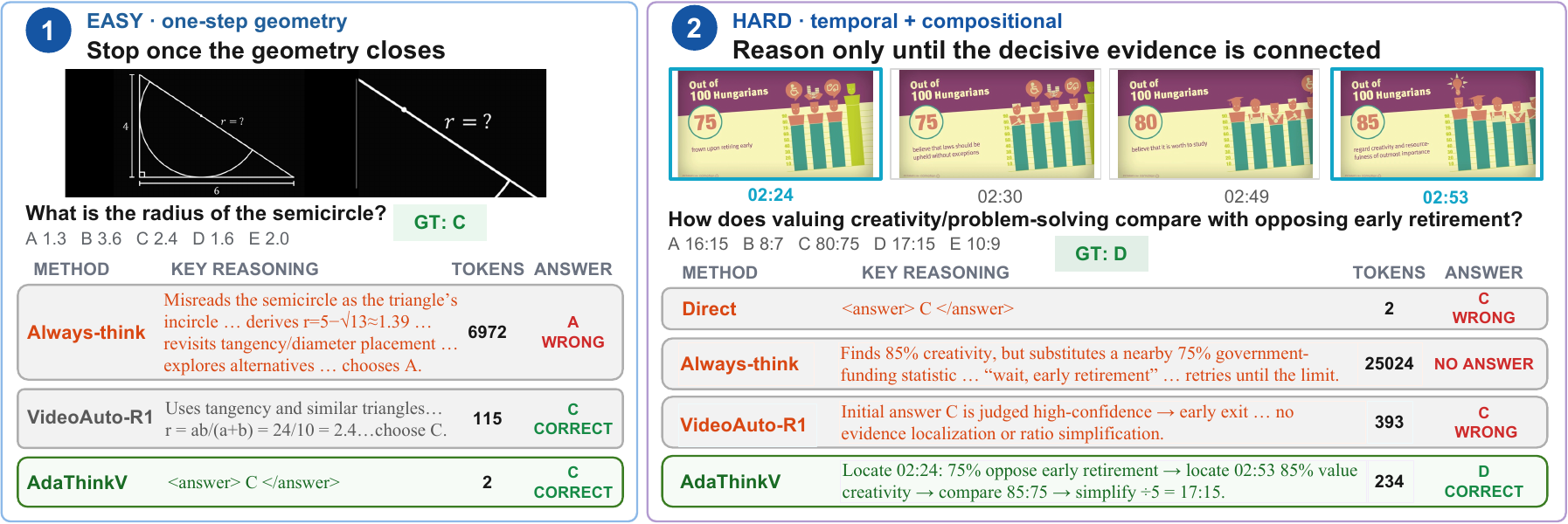}
\caption{Why does response mode selection matter? Excessive reasoning can overcomplicate simple questions, while confidence-routed VideoAuto-R1 may stop before collecting evidence needed for compositional reasoning. AdaThinkV selects direct or explicit reasoning and terminates autoregressively.}
\label{fig:teaser}
\end{figure*}

\section{Introduction}
Explicit chain-of-thought (CoT) can improve complex temporal reasoning in video multimodal large language models (MLLMs) \cite{wei2022chain,feng2025videor1}, yet not every question warrants a long visible trace. Simple perception can often be handled directly, whereas event ordering, state tracking, and causal analysis may require multi-step reasoning. Always reasoning wastes tokens and can overcomplicate simple questions, whereas always answering directly can underthink compositional ones. As \cref{fig:teaser} illustrates, a capable video MLLM should learn not only \emph{how} to reason, but also \emph{whether} explicit reasoning is worth its cost.

Adaptive reasoning methods switch between concise and explicit reasoning \cite{tu2025autothink,zhang2025adaptthink,fang2025thinkless}. For video, VideoAuto-R1 selects a response mode based on the confidence of a preliminary answer \cite{liu2026videoautor1}. Confidence, however, is only an indirect proxy: uncertainty does not imply that further reasoning will help, while high confidence can coexist with missed temporal evidence. The relevant question is not whether the model is uncertain, but whether additional reasoning improves the answer enough to justify its token cost. We formulate this quantity as the \emph{conditional net utility of reasoning}: the empirical accuracy gain from explicit reasoning minus its additional generation cost for the same input.

We introduce \textbf{AdaThinkV}, which learns to choose between explicit reasoning and direct answering without offline difficulty labels, confidence thresholds, or an external router. We first propose ThinkGain to estimate the prompt-specific utility of explicit reasoning from controlled \texttt{THINK}/\texttt{ANSWER} rollouts, accounting for both accuracy gain and additional response length. 
However, a fixed rollout group may not sufficiently explore difficult prompts, yielding only unsuccessful responses with little variation in accuracy rewards. Dynamic sampling addresses this issue by oversampling prompts and filtering out zero-variance groups to retain a fixed number of effective prompts, potentially biasing observed reward dispersion upward \cite{yu2025dapo}. To mitigate this bias, we introduce Variance Recovery Policy Optimization (VRPO), which progressively expands the same group while retaining existing samples for more reliable variance estimation. In inference, AdaThinkV generates the mode marker and response in one autoregressive sequence.

Across unified evaluation settings covering temporal reasoning, video mathematics, and scientific video understanding, AdaThinkV attains 40.79 mean accuracy with 257.20 output tokens. Relative to VideoAuto-R1-Qwen3, it improves accuracy by 2.98 points with 22.7\% fewer tokens; relative to Qwen3-VL-8B-Thinking, it gains 3.66 points with 95.8\% fewer tokens.

Our main contributions are fourfold:
\begin{itemize}
    \item We formulate adaptive video reasoning as prompt-level utility estimation and introduce AdaThinkV, which selects whether to reason explicitly without difficulty labels, confidence thresholds, or an external router.
    \item We introduce ThinkGain, a paired-rollout estimator of prompt-level reasoning utility that balances accuracy improvement against output length.
    \item We introduce VRPO, a complementary rollout allocation strategy that retains and expands groups with unsuccessful, low-dispersion rewards to recover informative RL signals from difficult prompts.
    \item Experiments across diverse video benchmarks show that AdaThinkV improves accuracy and efficiency over adaptive and reasoning baselines. Controlled ablations reveal that ThinkGain primarily improves the accuracy--token trade-off, whereas VRPO raises absolute accuracy.
\end{itemize}

%% file: sec/2_related_work.tex
\section{Related Work}

\noindent\textbf{Video reasoning and efficiency.}
Video reasoning has advanced through reinforcement learning, temporal grounding, structured distillation, and explicit reasoning traces \cite{feng2025videor1,lin2026visd,zhu2025memorize,wang2025timer1,tian2025ddavs,fei2025videoofthought,maaz2025videor2}. Efficiency-oriented methods shorten reasoning traces, compress visual inputs, or scale perception and test-time computation \cite{zhong2025rethinking,yan2025videochatr15,wang2025videorts,buch2025flexible,zhang2025vital}. These methods improve reasoning or reduce generation and perception costs, but most do not directly optimize whether visible reasoning is worthwhile for each question. AdaThinkV studies this cost-aware, per-question decision in video reasoning.

\noindent\textbf{Adaptive reasoning.}
AutoThink \cite{tu2025autothink} and AdaptThink \cite{zhang2025adaptthink} choose between direct and explicit reasoning, while Thinkless \cite{fang2025thinkless} separates mode control from response optimization. Switch-Reasoner uses fixed direct and thinking branches to construct sample-level supervision from their relative success and adds a global controller to balance mode usage \cite{fang2026switchreasoner}. Other methods control latent reasoning or the disclosure of visible traces \cite{li2026adaptivethinking,wei2026whentothink}. For video reasoning, VideoAuto-R1 selects a mode using the confidence of a preliminary answer \cite{liu2026videoautor1}. AdaThinkV instead estimates a length-aware utility from prompt-matched branches, applies a dead zone to uncertain comparisons, and separates marker supervision from conditional continuation learning. It generates the marker and response in one autoregressive sequence, requiring neither a preliminary answer nor an external router.

\noindent\textbf{Group relative RL and rollout allocation.}
GRPO optimizes a policy with group-relative advantages and no value model, while DAPO extends it with dynamic sampling that filters groups with no accuracy variation \cite{shao2024deepseekmath,yu2025dapo}. Later methods recover signals from such groups through modified credit assignment, advantage shaping, or reference-guided repair \cite{jin2026dgpo,le2025zerovariance,le2026sort}. Other work reduces rollout waste through prompt filtering \cite{zheng2025greso,li2025temporalrlt} or allocates rollouts using predicted statistics, pilot samples, and adaptive budgets \cite{nguyen2026adaptive,kim2026pilotcommit}. VRPO instead retains the current unsuccessful group and adds balanced mode rollouts until task success or sufficient accuracy reward dispersion emerges, recovering a learning signal without reference-guided repair or a learned prompt allocator.

%% file: sec/3_method.tex
\begin{figure*}[t]
\centering
\includegraphics[width=0.98\textwidth]{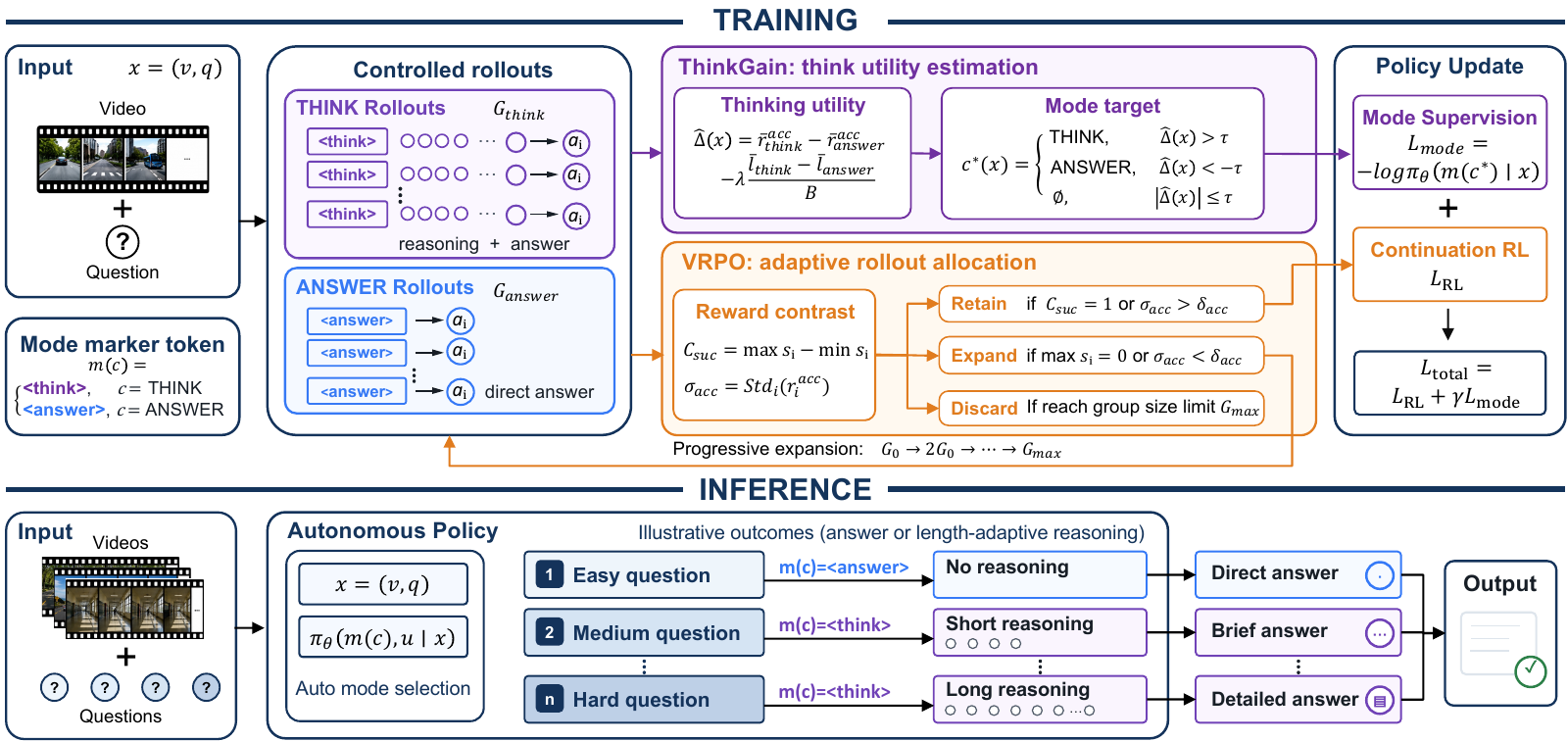}
\caption{AdaThinkV RL training and inference pipeline. ThinkGain constructs mode supervision from prompt-level  \texttt{THINK}/\texttt{ANSWER} branches. VRPO expands groups with weak reward contrast signals. 
At inference, the model autoregressively generates a mode marker followed by a variable-length continuation in a single sequence.
}
\label{fig:method_overview}
\end{figure*}

\section{Method}

\subsection{Overview}

Given a video $v$ and a question $q$, let $x=(v,q)$. AdaThinkV generates a response with an autoregressive policy. Let $c\in\{\mathtt{THINK},\mathtt{ANSWER}\}$ denote the response mode and let $m(c)$ denote its mode marker. The two response formats are
\[
\begin{aligned}
\mathtt{THINK}:&\quad
\texttt{<think>}\ h\ \texttt{</think>}\\[-0.2em]
&\quad \texttt{<answer>}\ a\ \texttt{</answer>},\\
\mathtt{ANSWER}:&\quad
\texttt{<answer>}\ a\ \texttt{</answer>},
\end{aligned}
\]
where $h$ is an explicit reasoning trace and $a$ is the final answer. Only the opening \texttt{<think>} or \texttt{<answer>} acts as the mode marker; the closing tags are format delimiters. All tokens after the marker form a continuation $u=(u_1,\ldots,u_T)$. A serialized mode marker may map to one or more tokenizer tokens. Writing $m(c)=(m_1,\ldots,m_{K_c})$, its probability is
\[
\pi_\theta(m(c)\mid x)
=\prod_{k=1}^{K_c}
\pi_\theta(m_k\mid x,m_{<k}),
\]
and the complete response factorizes as
\[
\pi_\theta(m(c),u\mid x)
=\pi_\theta(m(c)\mid x)
\prod_{t=1}^{T}\pi_\theta(u_t\mid x,m(c),u_{<t}).
\]

We first initialize the policy through an SFT cold start on balanced examples in both formats and then optimize it with reinforcement learning. Both modes must be explored to estimate when reasoning is useful, while group-relative optimization requires variation in accuracy rewards. As shown in \cref{fig:method_overview}, \textbf{ThinkGain} estimates prompt-level mode utility through controlled branches, whereas \textbf{Variance Recovery Policy Optimization (VRPO)} complements it by allocating additional rollouts to weak-signal groups. The shared training loop optimizes autonomous mode selection and conditional continuation generation.

\subsection{ThinkGain: Prompt-Matched Estimation of Reasoning Gain}

\noindent\textbf{Controlled rollouts.}
For a rollout group of size $G$, we force a fraction $\rho$ of responses to begin with the \texttt{THINK} mode marker and the remainder with the \texttt{ANSWER} mode marker:
\[
G_{\rm think}=\lfloor\rho G\rfloor,
\qquad
G_{\rm answer}=G-G_{\rm think}.
\]
Mode identity is assigned by rollout position rather than inferred from generated text. Because both branches use the same input and decoding configuration, their within-prompt comparison removes between-prompt difficulty variation while retaining finite-sample generation noise. It estimates conditional differences in accuracy and continuation length. When VRPO expands a group, ThinkGain is recomputed over all rollouts while maintaining balanced allocation.

\noindent\textbf{Mode utility with a length cost.}
For response $y_i$, let $r_i^{\rm acc}$ be its normalized task accuracy reward, where $0\leq r_i^{\rm acc}\leq1$, and let $\ell_i=|u_i|$ be its continuation length measured by the tokenizer after the opening mode marker. For mode $c$, define
\[
\overline{r}^{\rm acc}_{c}
=\frac{1}{G_c}\sum_{i:c_i=c}r_i^{\rm acc},
\qquad
\overline{\ell}_{c}
=\frac{1}{G_c}\sum_{i:c_i=c}\ell_i.
\]
ThinkGain estimates the net utility of explicit reasoning on prompt $x$ as
\begin{equation}
\label{eq:thinkgain}
\widehat{\Delta}(x)=
\overline{r}^{\rm acc}_{\rm think}
-\overline{r}^{\rm acc}_{\rm answer}
-\lambda\frac{\overline{\ell}_{\rm think}-\overline{\ell}_{\rm answer}}{B}.
\end{equation}
In \cref{eq:thinkgain}, $B$ normalizes length and $\lambda$ controls the extra decoding cost of \texttt{THINK} relative to \texttt{ANSWER}. A positive value indicates that the empirical accuracy gain from explicit reasoning outweighs its additional generation cost.

We use a margin $\tau\geq0$ to define a dead zone and convert this estimate into a mode target:
\begin{equation}
\label{eq:mode_target}
c^*(x)=
\begin{cases}
\mathtt{THINK}, & \widehat{\Delta}(x)>\tau,\\
\mathtt{ANSWER}, & \widehat{\Delta}(x)<-\tau,\\
\emptyset, & |\widehat{\Delta}(x)|\leq\tau.
\end{cases}
\end{equation}
Because $\widehat{\Delta}(x)$ is a Monte Carlo estimate based on finitely many samples, $c^*(x)$ is a utility preference rather than a ground-truth difficulty label. The dead zone suppresses small estimated differences but does not guarantee statistical certainty. We therefore treat the supervision as noisy. Additional ablations and diagnostic analyses are provided in the supplement.

\noindent\textbf{Mode selection learning.}
Forced mode markers define a stratified intervention rather than actions sampled from the old mode policy. We mask these tokens from the RL loss, so only the continuation is optimized conditional on the assigned mode. Autonomous mode selection is trained toward the target in \cref{eq:mode_target} with a separate supervised likelihood evaluated before the mode marker:
\begin{equation}
\label{eq:mode_loss}
\mathcal{L}_{\rm mode}
=-\frac{1}{N_{\rm mode}}
\sum_{\substack{x\in\mathcal{B}_{\rm RL}\\c^*(x)\neq\emptyset}}
\log\pi_\theta(m(c^*(x))\mid x),
\end{equation}
where $N_{\rm mode}$ is the number of prompts in the sum. We set this term to zero when $N_{\rm mode}=0$. Thus, marker positions receive no direct policy-gradient loss; their explicit token-level supervision is provided by \cref{eq:mode_loss}. Continuation updates can still affect marker probabilities indirectly through the shared policy parameters.

\begin{figure}[t]
\centering
\includegraphics[width=\columnwidth]{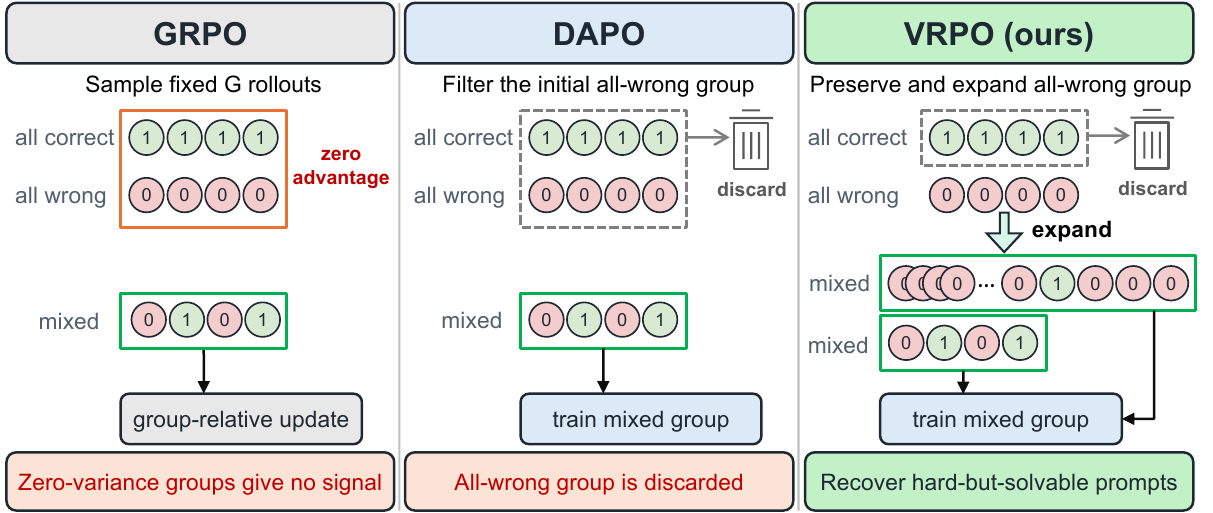}
\caption{Rollout allocation comparison. GRPO uses fixed groups, DAPO replaces zero-contrast prompts, and VRPO retains and expands unsuccessful groups with low accuracy-reward dispersion.}
\label{fig:rollout_comparison}
\end{figure}

\subsection{Variance Recovery Policy Optimization}

VRPO follows DAPO's policy optimization design but replaces its dynamic sampling rule \cite{yu2025dapo}. Rather than replacing a prompt with a weak learning signal and discarding its trajectories, VRPO preserves the current group and progressively adds rollouts. \Cref{fig:rollout_comparison} contrasts this allocation rule with GRPO and DAPO.

\noindent\textbf{Composite trajectory reward.}
We optimize retained trajectories with the total reward $r_i$ as
\begin{equation}
\label{eq:trajectory_reward}
\begin{aligned}
r_i&=r_i^{\rm acc}+r_i^{\rm fmt}+r_i^{\rm ol}+g_i,\\
g_i&=\alpha\,\mathbf{1}[c_i=c^*(x)].
\end{aligned}
\end{equation}
In \cref{eq:trajectory_reward}, $r_i^{\rm fmt}$ scores compliance with the assigned serialization, and $r_i^{\rm ol}$ follows DAPO's soft penalty near the length limit. DAPO's hard mask removes completions truncated at this limit from the policy gradient loss. Since $c_i\in\{\mathtt{THINK},\mathtt{ANSWER}\}$, no mode-consistency bonus is assigned when $c^*(x)=\emptyset$. Task correctness and VRPO allocation depend only on $r_i^{\rm acc}$, so $g_i$ supplements rather than replaces the accuracy reward. Because forced marker tokens are masked, $g_i$ changes the advantages of continuation tokens but not the probability of the target marker. In contrast, $\mathcal{L}_{\rm mode}$ acts directly on the autonomous marker probability. The two terms use the same target but optimize different conditional factors of the response.

\noindent\textbf{Task success and reward dispersion.}
VRPO distinguishes a useful accuracy signal from task success. These concepts coincide for binary rewards based on exact answers but differ for continuous rewards such as temporal IoU. We define the binary accuracy $s_i$ as
\[
s_i=
\begin{cases}
r_i^{\rm acc},
& \text{for binary rewards},\\
\mathbf{1}[r_i^{\rm acc}\geq\eta_{\rm task}],
& \text{for continuous rewards}.
\end{cases}
\]
For continuous tasks, $\eta_{\rm task}$ is fixed by the task success criterion rather than tuned on downstream test sets.

For the current group $\mathcal{Y}_x=\{y_i\}_{i=1}^{G_x}$, let
\[
\overline r_x^{\rm acc}
=\frac{1}{G_x}\sum_{i=1}^{G_x}r_i^{\rm acc},
\qquad
C_{\rm suc}(x)=\max_i s_i-\min_i s_i,
\]
and define its empirical accuracy reward dispersion as
\[
\sigma_{\rm acc}(x)
=
\sqrt{
\frac{1}{G_x}
\sum_{i=1}^{G_x}
\left(r_i^{\rm acc}-\overline r_x^{\rm acc}\right)^2
}.
\]
A group provides a useful learning signal if it contains either contrast between successful and unsuccessful responses, $C_{\rm suc}(x)=1$, or sufficient accuracy reward dispersion, $\sigma_{\rm acc}(x)>\delta_{\rm acc}$. VRPO expands only groups in which all current responses are unsuccessful and their accuracy rewards remain insufficiently differentiated:
\begin{equation}
\label{eq:vrpo_action}
\operatorname{action}(x)=
\begin{cases}
\mathrm{retain},
& \substack{C_{\rm suc}(x)=1\ \lor\\
\sigma_{\rm acc}(x)>\delta_{\rm acc}},\\
\mathrm{expand},
& \substack{\max_i s_i=0,\ 
\sigma_{\rm acc}(x)\leq\delta_{\rm acc},\\
G_x<G_{\max}},\\
\mathrm{discard}, & \text{otherwise}
\end{cases}.
\end{equation}
For continuous rewards, \cref{eq:vrpo_action} retains a group in which all responses are successful when $\sigma_{\rm acc}(x)>\delta_{\rm acc}$ because its graded rewards still support relative learning. Such a group is discarded when its dispersion is insufficient. Each expansion preserves existing trajectories and the controlled mode ratio, and only retained groups contribute to $\mathcal{B}_{\rm RL}$.

Expansion stops according to rewards, and the retained group is reused to estimate the ThinkGain target and compute the policy update. This reuse may introduce selection bias from optional stopping. We quantify the effect with diagnostics for each expansion stage and a comparison that uses an independently resampled update group in the supplement.

\begin{figure*}[t]
\centering
\includegraphics[width=\textwidth]{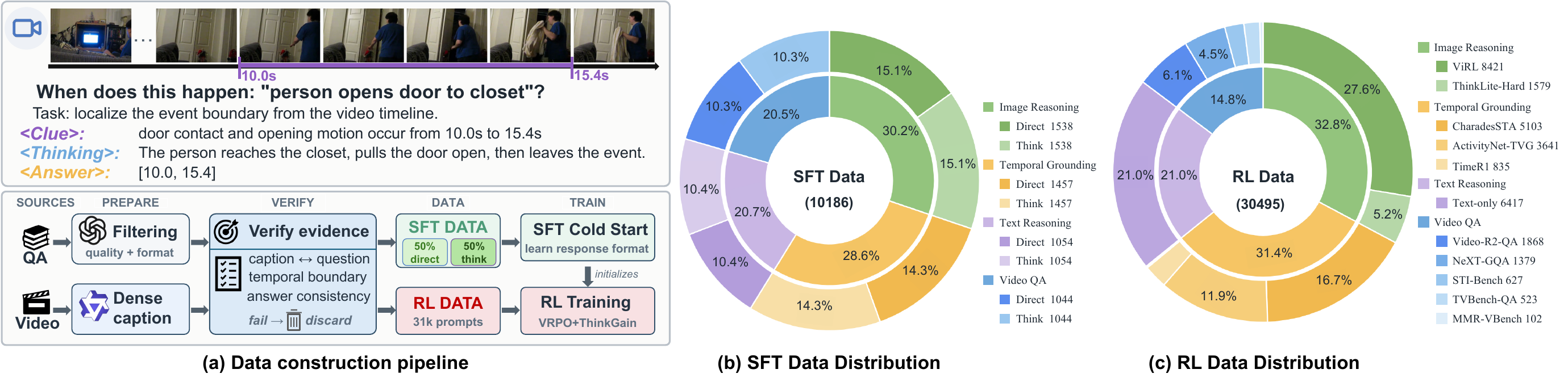}
\caption{Training data construction pipeline and data distribution.}
\label{fig:data_full}
\end{figure*}

\input{Tab/main_results}

\noindent\textbf{Optimization relative to each group.}
For each retained group, the trajectory advantage $\widehat{A}_i$ is obtained by normalizing the composite rewards across all $G_x$ rollouts:

\[
\widehat{A}_{i}=
\frac{r_i-\operatorname{mean}(\{r_j\}_{j=1}^{G_x})}
{\operatorname{std}(\{r_j\}_{j=1}^{G_x})+\epsilon}.
\]
Here, $\epsilon>0$ is a numerical stabilizer. The same trajectory advantage is applied to every continuation token in response $i$. For continuation token $u_{i,t}$, we define the importance ratio $s_{i,t}(\theta)$ as
\[
s_{i,t}(\theta)=
\frac{\pi_\theta(u_{i,t}\mid x,m(c_i),u_{i,<t})}
{\pi_{\theta_{\rm old}}(u_{i,t}\mid x,m(c_i),u_{i,<t})},
\]
\[
\widetilde{s}_{i,t}(\theta)=
\operatorname{clip}\!\left(
s_{i,t}(\theta),1-\epsilon_{\rm low},1+\epsilon_{\rm high}
\right).
\]
We use $\epsilon_{\rm high}>\epsilon_{\rm low}$. The tighter lower clipping bound limits destructive probability decreases, whereas the looser upper bound allows advantageous but initially rare tokens to gain probability more quickly. Let $\mu_i$ be DAPO's binary hard overlong mask. It equals zero only when response $i$ is truncated at the generated token limit. Following the clipped policy objective of PPO and the token aggregation of DAPO \cite{schulman2017ppo,yu2025dapo}, the optimization objective is
\begin{equation}
\label{eq:rl_loss}
\begin{aligned}
\mathcal{L}_{\rm RL}
&=-\frac{1}{|\mathcal{B}_{\rm RL}|}
\sum_{x\in\mathcal{B}_{\rm RL}}
\frac{1}{\sum_{i=1}^{G_x}\mu_i|u_i|}
\sum_{i=1}^{G_x}\mu_i\sum_{t=1}^{|u_i|}
\\[-0.2em]
&\quad
\min\!\left(
s_{i,t}(\theta)\widehat{A}_{i},
\widetilde{s}_{i,t}(\theta)\widehat{A}_{i}
\right).
\end{aligned}
\end{equation}
In \cref{eq:rl_loss}, any prompt with zero unmasked continuation tokens is removed from $\mathcal{B}_{\rm RL}$, and $\mathcal{L}_{\rm RL}=0$ if the buffer is empty. This normalization gives every unmasked continuation token equal weight within its prompt. It avoids the implicit $1/|u_i|$ weighting that arises when each response is averaged separately, and it prevents larger recovered groups from dominating the minibatch. The complete objective is
\begin{equation}
\label{eq:total_objective}
\mathcal{L}_{\rm total}
=\mathcal{L}_{\rm RL}+\gamma\mathcal{L}_{\rm mode}.
\end{equation}
In \cref{eq:total_objective}, $\gamma$ controls mode supervision. As in DAPO, we use no KL penalty or separate reference model. We retain $\pi_{\theta_{\rm old}}$ only to compute policy optimization ratios. Continuation tokens are sampled from $\pi_{\theta_{\rm old}}$ conditional on the forced marker, so their ratios remain on policy for that conditional distribution. At test time, mode forcing is removed and the opening marker is generated autoregressively.

%% file: Tab/main_results.tex
\begin{table*}[t]
\centering
\small
\setlength{\tabcolsep}{2.6pt}
\begin{tabular}{@{}llcccccccc@{}}
\toprule
Model & Mode & VRB & VMath & VMath-M & MMRV & MMRV-CoT & SciV & Avg. & Tok. \\
\midrule
Qwen2.5-VL-7B        & Direct & 3.54 & 29.76 & 55.95 & 42.00 & 34.77 & 27.90 & 32.32 & 400.82 \\
Qwen3-VL-8B-Instruct & Direct & 5.56 & 30.71 & 59.17 & 45.58 & 41.21 & 33.10 & 35.89 & 524.62 \\
Qwen3-VL-8B-Thinking & Think only & 6.11 & 38.81 & 62.38 & 42.72 & 42.24 & 30.50 & 37.13 & 6184.98 \\
Temporal-RLT        & Think only & 2.64 & 29.52 & 55.77 & 41.29 & 35.96 & 26.80 & 32.00 & 343.20 \\
Time-R1-7B          & Think only & 2.99 & 29.52 & 56.31 & 42.16 & 36.12 & 27.60 & 32.45 & 307.68 \\
VideoRFT            & Think only & 2.22 & 25.71 & 54.94 & 40.89 & 40.49 & 26.00 & 31.71 & 427.35 \\
Video-R1-7B         & Think only & 1.67 & 22.62 & 53.87 & 39.94 & 34.77 & 24.70 & 29.59 & 125.03 \\
Video-RTS           & Think only & 3.13 & 28.33 & 56.31 & 42.24 & 37.07 & 28.40 & 32.58 & 483.78 \\
LOVE-R1             & Think only & 2.50 & 18.57 & 49.94 & 41.21 & 40.02 & 14.10 & 27.72 & 848.18 \\
Video-R2            & Think only & 1.81 & 29.29 & 56.55 & 39.62 & 35.72 & 27.70 & 31.78 & 346.05 \\
VideoChat-R1.5      & Think only & 2.78 & 29.05 & 55.95 & 43.68 & 40.65 & 28.00 & 33.35 & 251.22 \\
VideoAuto-R1-Qwen2.5 & Auto & 3.33 & 30.95 & 56.07 & 49.96 & 42.08 & 31.10 & 35.58 & 268.37 \\
VideoAuto-R1-Qwen3  & Auto & 4.93 & 31.43 & 58.93 & 50.84 & 47.81 & 32.90 & 37.81 & 332.93 \\
\midrule
AdaThinkV w/o VRPO & Auto & 6.94 & 36.67 & 59.76 & 46.62 & 45.51 & 34.40 & 38.32 & 235.73 \\
AdaThinkV & Auto & \textbf{7.57} & \textbf{39.05} & \textbf{62.56} & \textbf{50.99} & \textbf{48.05} & \textbf{36.50} & \textbf{40.79} & 257.20 \\
\bottomrule
\end{tabular}
\caption{Comparison on video reasoning benchmarks. VRB, VMath, MMRV, and SciV denote VideoReasonBench, VideoMathQA, MMR-VBench, and SciVideoBench. Avg. is the average accuracy. Tok. is the average output token length.}
\label{tab:main_results}
\end{table*}

%% file: sec/4_experiments.tex
\begin{figure}[!t]
\centering
\includegraphics[width=1\columnwidth]{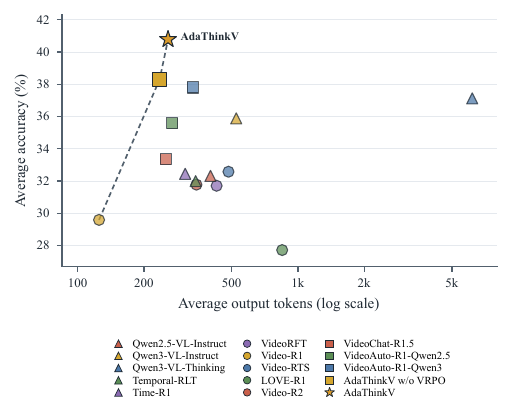}
\caption{Accuracy--token trade-off. AdaThinkV advances the empirical Pareto frontier, achieving higher average accuracy with fewer generated tokens.}
\label{fig:accuracy_token_pareto}
\end{figure}

\section{Experiments}
\subsection{Experimental Setup}
\noindent\textbf{Training data.}
\Cref{fig:data_full} summarizes our SFT cold-start set of 10186 traces and RL set of 30495 prompt-only inputs, spanning image reasoning, temporal grounding, text reasoning, and video QA. Within each task family, the SFT set is balanced between direct-answer and explicit-reasoning traces. The RL set does not contain prefabricated mode-specific responses; instead, \texttt{THINK} and \texttt{ANSWER} exploration is balanced online through controlled rollout allocation.  Data construction and filtering details are provided in the supplement.

\input{Tab/general_video_results}
\input{Tab/temporal_grounded_results}
\input{Tab/streaming_results}
\input{Tab/data_mixture}
\input{Tab/main_ablation}
\input{Tab/mode_ablation}
\input{Tab/vrpo}

\noindent\textbf{Evaluation protocol.}
We evaluate every model in \cref{tab:main_results} with a unified pipeline. Unless stated otherwise, we sample between 2 and 256 frames at 2 FPS, impose a visual budget of 4096 tokens, and allow up to 32768 generated tokens subject to the context window. StreamingBench and OVO-Bench follow a query-time-causal protocol: for a query issued at timestamp $t_q$, frames are sampled only from the visible prefix $[0,t_q]$, and no post-query frames are provided to the model. All models share inputs, parsers, and scoring rules while retaining their native templates. Component ablations share initialization, training data, and optimization. VideoReasonBench uses its semantic judge, whereas the remaining benchmarks use exact accuracy. Benchmark-specific settings and a generation-cap sensitivity analysis are provided in the supplement.

\noindent\textbf{Implementation details.}
Using Qwen3-VL-8B \cite{bai2025qwen3vl} as the policy backbone, we train the SFT and RL stages for one epoch each, with a frozen visual encoder and a global batch size of 32. AdamW uses a learning rate of $1\times10^{-6}$, weight decay $0.01$, and gradient clipping at $1.0$. Rollouts use temperature $1.0$, nucleus sampling with $p=0.95$, top $k=20$, and at most 4096 generated tokens. The clipping bounds are $\epsilon_{\rm low}=0.20$ and $\epsilon_{\rm high}=0.28$. ThinkGain uses $\rho=0.5$, $\lambda=0.1$, $B=512$, $\tau=0.05$, $\alpha=0.3$, and $\gamma=0.1$. VRPO starts from $G_0=8$, adds four rollouts per mode at each stage, and stops at $G_{\max}=32$. For continuous rewards, it uses $\eta_{\rm task}=0.5$ and $\delta_{\rm acc}=0.05$, neither of which is tuned on downstream test sets. Under the Qwen3-VL tokenizer, the opening \texttt{<think>} marker is one token and \texttt{<answer>} is three tokens. We use the full-sequence marker likelihood in \cref{eq:mode_loss}, without per-token length normalization, and mask every forced marker token from the continuation loss. Training is distributed across two eight-GPU nodes, for 16 NVIDIA H100 GPUs in total, with bfloat16 optimization, DeepSpeed ZeRO-1, colocated vLLM generation, and seed 42. Training cost is reported in \cref{tab:vrpo}, and detailed infrastructure is provided in the supplement.

\subsection{Main Results}

\noindent\textbf{Main video reasoning results.}
\Cref{tab:main_results} reports results on VideoReasonBench, VideoMathQA, MMR-VBench, and SciVideoBench \cite{liu2025videoreasonbench,rasheed2025videomathqa,zhu2025mmrv,deng2025scivideobench}. We compare general-purpose, specialized, and adaptive reasoning models \cite{bai2025qwen25vl,bai2025qwen3vl,li2025temporalrlt,wang2025timer1,wang2025videorft,feng2025videor1,wang2025videorts,fu2025lover1,maaz2025videor2,yan2025videochatr15,liu2026videoautor1}. Under the unified protocol, AdaThinkV reaches an average accuracy of 40.79, exceeding VideoAuto-R1-Qwen3 by 2.98 points with 22.7\% fewer generated tokens; the paired-bootstrap 95\% confidence interval for this gain is $[2.21,3.75]$. As shown in \cref{fig:accuracy_token_pareto}, both AdaThinkV variants lie on the empirical accuracy--token Pareto frontier. The average across the four canonical benchmarks is 33.53, while the average across benchmark families is 36.10.

\noindent\textbf{General video understanding.}
\Cref{tab:general_video} extends the comparison to Video-MME \cite{fu2024videomme}, MVBench \cite{li2023mvbench}, LongVideoBench \cite{wu2024longvideobench}, MMVU \cite{zhao2025mmvu}, and Video-MMMU \cite{hu2025videommmu}. Against the broader set of baselines, including VITAL and LongVILA-R1 \cite{zhang2025vital,chen2025longvilar1}, AdaThinkV ranks first on four of the five benchmarks and remains within 0.4 points of the best Video-MME result.

\noindent\textbf{Temporal grounding and streaming video.}
\Cref{tab:temporal_grounded_results} evaluates Charades-STA \cite{gao2017tall}, ActivityNet Captions \cite{krishna2017densecaptioning}, and NExT-GQA \cite{xiao2024nextgqa}, while \cref{tab:streaming} covers StreamingBench and OVO-Bench under the query-time-causal visible-prefix protocol \cite{lin2024streamingbench,li2025ovobench}. Across the established temporal and streaming baselines \cite{ren2023timechat,zeng2024timesuite,chen2024timemarker,huang2025ovbench,zeng2025streamforest,zhang2026hermes}, AdaThinkV ranks first on all but one temporal-grounding metric and on both streaming benchmarks.

\subsection{Ablation Study}

\noindent\textbf{SFT cold start and RL stages.}
\Cref{tab:data_mixture} shows that adding RL after the balanced SFT cold start improves mean accuracy by 3.99 points while maintaining over 99\% format compliance.

\noindent\textbf{ThinkGain and VRPO.}
\Cref{tab:main_ablation} compares GRPO, SAPO, and DAPO \cite{shao2024deepseekmath,gao2026sapo,yu2025dapo}. ThinkGain improves DAPO by 2.62 points while more than halving output length; on VRPO, it adds 0.30 points while reducing length by 46.6\%. Thus, ThinkGain primarily improves the accuracy--decoded-length trade-off, whereas VRPO adds 2.37 points with ThinkGain fixed. The gain is consistent across three seeds; further decompositions are reported in the supplement.

\noindent\textbf{Inference modes.}
\Cref{tab:mode_ablation} shows that automatic inference is 0.71 points below forced \texttt{THINK} (95\% CI $[-1.57,0.15]$) with 62.9\% fewer tokens, and 2.11 points above forced \texttt{ANSWER}. Across difficulty tertiles, the thinking rate rises from 25.6\% to 69.1\% and the mean \texttt{THINK} length from 151 to 612 tokens; prompt-level reasoning length is moderately correlated with difficulty ($\rho=0.39$).



\noindent\textbf{Rollout allocation.}
\Cref{tab:vrpo} shows that increasing fixed $G$ from 8 to 32 nearly quadruples training cost for only a 0.87-point gain. At comparable compute, VRPO reaches 27.71 accuracy with 716 GPU-hours, compared with 26.25 at 718 GPU-hours for the fixed-allocation control.

%% file: Tab/general_video_results.tex
\begin{table}[!t]
\centering
\footnotesize
\setlength{\tabcolsep}{1pt}
\begin{tabular}{@{}lccccc@{}}
\toprule
Model & V-MME & MVB & LongVB & MMVU & V-MMMU \\
\midrule
Qwen2.5-VL-7B & 66.0 & 67.1 & 60.9 & 66.2 & 54.7 \\
Qwen3-VL-8B & \textbf{72.5} & 69.4 & 67.6 & 69.9 & 61.0 \\
Temporal-RLT & 57.6 & 68.1 & -- & 65.0 & -- \\
VideoRFT & 59.8 & 62.1 & -- & 68.5 & 51.1 \\
Video-R1-7B & 61.8 & 65.5 & -- & 65.0 & 51.4 \\
Video-RTS & 63.0 & -- & 56.6 & 66.4 & 52.7 \\
VITAL & 64.1 & -- & -- & 68.7 & 54.2 \\
LongVILA-R1-7B & 65.1 & 67.6 & 58.0 & -- & -- \\
LOVE-R1 & 66.2 & 66.6 & 60.1 & -- & -- \\
VideoChat-R1.5 & 65.2 & 70.6 & 61.4 & -- & 49.6 \\
VideoAuto-R1-Q2.5 & 67.3 & 71.0 & 60.5 & 69.7 & 58.6 \\
VideoAuto-R1-Q3 & 71.7 & 72.0 & 67.4 & 71.1 & 65.0 \\
AdaThinkV & 72.1 & \textbf{72.5} & \textbf{68.1} & \textbf{71.4} & \textbf{65.3} \\
\bottomrule
\end{tabular}
\caption{General video understanding benchmarks. V-MME, MVB, LongVB, and V-MMMU abbreviate Video-MME, MVBench, LongVideoBench, and Video-MMMU; Q2.5 and Q3 indicate the Qwen backbones.}
\label{tab:general_video}
\end{table}

%% file: Tab/temporal_grounded_results.tex
\begin{table}[t]
\centering
\footnotesize
\setlength{\tabcolsep}{0.7pt}
\begin{tabular}{lcccccc}
\toprule
 & \multicolumn{2}{c}{Charades-STA} & \multicolumn{2}{c}{ActivityNet} & \multicolumn{2}{c}{NExT-GQA} \\
\cmidrule(lr){2-3}\cmidrule(lr){4-5}\cmidrule(lr){6-7}
Model & R@0.5 & mIoU & R@0.5 & mIoU & Acc. & mIoU \\
\midrule
Qwen2.5-VL-7B & 59.6 & 52.9 & 22.6 & 26.9 & 53.3 & 20.2 \\
TimeChat & 27.5 & 31.2 & 27.8 & 30.4 & 28.8 & 17.4 \\
TimeSuite & 67.1 & -- & -- & -- & -- & -- \\
TimeMarker & 51.9 & 48.4 & 50.7 & 49.5 & -- & -- \\
Temporal-RLT & 67.9 & 57.0 & 38.4 & 39.0 & 78.7 & 37.3 \\
Time-R1 & 72.2 & 58.8 & \textbf{55.6} & 52.1 & -- & -- \\
VITAL & 72.0 & 59.9 & 50.8 & 49.8 & 78.7 & 43.0 \\
VideoChat-R1.5 & 71.6 & 60.6 & 32.3 & 35.3 & -- & -- \\
VideoAuto-R1-Q2.5 & 70.8 & 60.0 & 48.5 & 47.6 & 80.6 & 36.7 \\
VideoAuto-R1-Q3 & 74.9 & 63.7 & 54.3 & 51.9 & 81.1 & 44.2 \\
\midrule
AdaThinkV & \textbf{75.2} & \textbf{64.0} & 54.9 & \textbf{52.2} & \textbf{81.4} & \textbf{44.5} \\
\bottomrule
\end{tabular}
\caption{Temporal grounding benchmarks. Localization is measured by R@0.5 and mIoU; NExT-GQA also reports answer accuracy.}
\label{tab:temporal_grounded_results}
\end{table}

%% file: Tab/streaming_results.tex
\begin{table}[t]
\centering
\footnotesize
\setlength{\tabcolsep}{4.0pt}
\begin{tabular}{@{}lcc@{}}
\toprule
Model & StreamingBench & OVO-Bench \\
\midrule
Qwen2.5-VL-7B & 73.31 & 52.28 \\
TimeChat-Online-7B & 75.28 & 51.80 \\
StreamForest-7B & 77.26 & 56.60 \\
HERMES-7B & 79.44 & 59.20 \\
Qwen3-VL-8B & 78.65 & 63.51 \\
\midrule
AdaThinkV & \textbf{79.79} & \textbf{64.02} \\
\bottomrule
\end{tabular}
\caption{Streaming video understanding. StreamingBench reports RTVU accuracy; OVO-Bench averages its Real-Time and Backward tracks.}
\label{tab:streaming}
\end{table}

%% file: Tab/data_mixture.tex
\begin{table}[t]
\centering
\footnotesize
\setlength{\tabcolsep}{3.0pt}
\begin{tabular}{@{}lcccc@{}}
\toprule
Training recipe & Acc. & Fmt. & Tok. & Think \\
\midrule
RL only & 35.59 & 92.4 & 660.20 & 87.5 \\
SFT only & 36.80 & 99.1 & 260.48 & 33.9 \\
SFT $\rightarrow$ RL & 40.79 & 99.3 & 257.20 & 49.1 \\
\bottomrule
\end{tabular}
\caption{SFT and RL stage ablation. Acc. and Tok. use the six-setting evaluation in \cref{tab:main_results}; Fmt. and Think are percentages over the same examples.}
\label{tab:data_mixture}
\end{table}

%% file: Tab/main_ablation.tex
\begin{table}[t]
\centering
\small
\setlength{\tabcolsep}{2pt}
\begin{tabular}{@{}cccccccc@{}}
\toprule
Method & ThinkGain & VRB & VMath & MMRV & SciV & Avg. & Tok. \\
\midrule
GRPO & No & 5.42 & 30.24 & 42.35 & 31.80 & 27.45 & 612.40 \\
SAPO & No & 5.76 & 31.19 & 43.10 & 32.20 & 28.06 & 568.60 \\
DAPO & No & 6.04 & 31.67 & 43.85 & 32.60 & 28.54 & 524.25 \\
VRPO & No & 7.43 & 38.57 & 50.55 & 36.35 & 33.23 & 481.85 \\
DAPO & Yes & 6.94 & 36.67 & 46.62 & 34.40 & 31.16 & 235.73 \\
VRPO & Yes & 7.57 & 39.05 & 50.99 & 36.50 & 33.53 & 257.20 \\
\bottomrule
\end{tabular}
\caption{ThinkGain and VRPO ablation. Results average VRB, VMath, MMRV, and SciV.}
\label{tab:main_ablation}
\end{table}

%% file: Tab/mode_ablation.tex
\begin{table}[t]
\centering
\small
\setlength{\tabcolsep}{4.0pt}
\begin{tabular}{@{}llccc@{}}
\toprule
Policy & Mode & Think & Tok. & Avg. \\
\midrule
VideoAuto-R1 & Direct & 0.0 & 15.60 & 28.82 \\
VideoAuto-R1 & Think & 100.0 & 2380.40 & 31.98 \\
VideoAuto-R1 & Auto & 16.8 & 370.43 & 30.03 \\
\midrule
AdaThinkV & Direct & 0.0 & 20.50 & 31.42 \\
AdaThinkV & Think & 100.0 & 693.95 & 34.24 \\
AdaThinkV & Auto & 47.2 & 257.20 & 33.53 \\
\bottomrule
\end{tabular}
\caption{Inference mode comparison. Avg. and Tok. average VRB, VMath, MMRV, and SciV; Think is the percentage of \texttt{THINK} openings.}
\label{tab:mode_ablation}
\end{table}

%% file: Tab/vrpo.tex

\begin{table}[!t]
\centering
\small
\setlength{\tabcolsep}{1.5pt}
\begin{tabular}{@{}lccccc@{}}
\toprule
Setting & Avg. $G$ & Seq. (M) & Tok. (M) & GPU-h & Acc. \\
\midrule
Fixed $G=8$ (DAPO) & 8.0 & 0.244 & 62.75 & 494 & 26.00 \\
Fixed $G=16$ & 16.0 & 0.488 & 125.49 & 988 & 26.52 \\
Fixed $G=32$ & 32.0 & 0.976 & 250.99 & 1975 & 26.87 \\
Fixed $G{\in}\{8,16\}$ mix & 11.6 & 0.354 & 91.00 & 718 & 26.25 \\
Expand all zero-contrast & 12.4 & 0.378 & 97.26 & 765 & 27.01 \\
VRPO & 11.6 & 0.354 & 90.98 & 716 & 27.71 \\
\bottomrule
\end{tabular}
  \caption{Training efficiency comparison between VRPO and other rollout settings. Accuracy is averaged over VRB, VMath, and SciV. The fixed mix is compute-matched to VRPO, whereas ``Expand all zero-contrast'' expands both all-wrong and all-correct groups.}
\label{tab:vrpo}
\end{table}

%% file: sec/6_conclusion.tex
\section{Conclusion}
We presented AdaThinkV, an adaptive framework for video reasoning that learns when explicit reasoning is worth its generation cost. ThinkGain estimates prompt-level reasoning utility by balancing accuracy improvement against additional output length, thereby supervising both mode selection and conditional response generation. VRPO retains and progressively expands all-unsuccessful, low-dispersion rollout groups, recovering informative learning signals from difficult yet solvable prompts. At inference, AdaThinkV generates the selected mode marker and response in a single autoregressive sequence,
without an external router or a preliminary answer. Under the unified evaluation protocol, AdaThinkV outperforms the strongest evaluated adaptive baseline by 2.98 accuracy points while generating 22.7\% fewer output tokens. Future work will extend this framework to image reasoning and other multimodal tasks and generalize its utility estimation and rollout allocation beyond scalar accuracy rewards to feedback from generative and rubric-based reward models.